\documentclass{article}
\usepackage{ijcai24}

\usepackage{times}
\usepackage{soul}
\usepackage{url}
\usepackage[hidelinks]{hyperref}
\usepackage[utf8]{inputenc}
\usepackage[small]{caption}
\usepackage{graphicx}
\usepackage{amsmath}
\usepackage{amsthm}
\usepackage{booktabs}
\usepackage{algorithm}
\usepackage{algorithmic}
\usepackage[switch]{lineno}
\usepackage{booktabs}
\usepackage{amssymb}
\usepackage{bbding}
\usepackage{multirow}

\title{Enhancing DETR's Variants through Improved Content Query and Similar Query Aggregation}

\author{
Yingying Zhang
\and
Chuangji Shi\and
Xin Guo \and
Jiangwei Lao \and 
Jian Wang \and
Jiaotuan Wang \and
Jingdong Chen
\affiliations
Ant Group
\emails
\{qichu.zyy, chuanji.scj, wenshuo.ljw,bobblair.wj,jingdongchen.cjd\}@antgroup.com
}

\begin{document}

\maketitle

\begin{abstract}

The design of the query is crucial for the performance of DETR and its variants. Each query consists of two components: a content part and a positional one. Traditionally, the content query is initialized with a zero or learnable embedding, lacking essential content information and resulting in sub-optimal performance. In this paper, we introduce a novel plug-and-play module, Self-Adaptive Content Query (SACQ), to address this limitation. The SACQ module utilizes features from the transformer encoder to generate content queries via self-attention pooling. This allows candidate queries to adapt to the input image, resulting in a more comprehensive content prior and better focus on target objects. However, this improved concentration poses a challenge for the training process that utilizes the Hungarian matching, which selects only a single candidate and suppresses other similar ones. To overcome this, we propose a query aggregation strategy to cooperate with SACQ. It merges similar predicted candidates from different queries, easing the optimization. Our extensive experiments on the COCO dataset demonstrate the effectiveness of our proposed approaches across six different DETR's variants with multiple configurations, achieving an average improvement of over 1.0 AP.

\end{abstract}

\section{Introduction}
\label{intro}

\begin{figure}[!h]
  \centering
  \includegraphics[width=1.\linewidth]{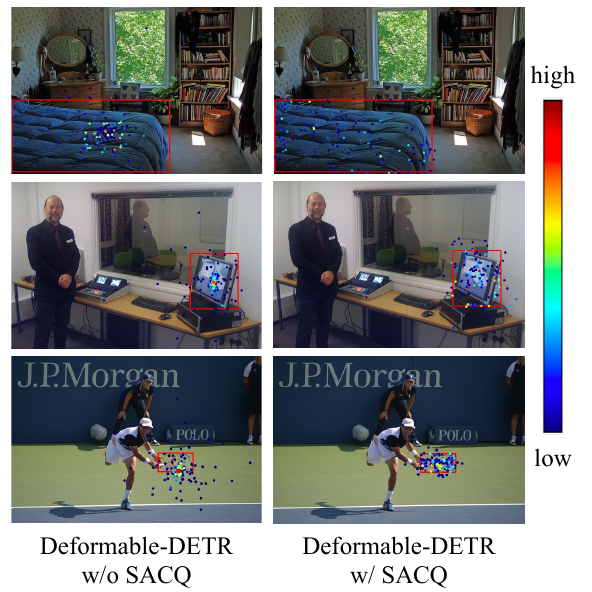}
  \caption{Comparison of multi-scale deformable attention of the first decoder layer between vanilla Deformable-DETR and Deformable-DETR with SACQ. We draw the sampling points and attention weights from multi-scale feature maps in one picture. Each sampling point is marked as a filled circle whose color indicates its attention weight. The red rectangle is the predicted bounding box of correspoinding query.}
   \label{REF}
\end{figure} 

Object detection is essential in various applications, such as autonomous driving, video surveillance, and robotic manipulation. Over the past few decades, convolutional architectures have driven significant advancements in detection methods \cite{Girshick_2015_ICCV,ren2015faster,Tian_2019_ICCV,he2017mask,Redmon_2016_CVPR,liu2016ssd,fpn}. These algorithms typically require a hand-designed module to generate anchors, which serve as preliminary candidates for object detection. Furthermore, non-maximum suppression (NMS) \cite{nms} is indispensable for preventing duplicate detections. Recently, Carion \textit{et al.} proposed a fully end-to-end object detection approach named DEtection TRansfomer (DETR) \cite{carion2020end}. In contrast to previous detection algorithms, DETR employs learned queries to predict objects uniquely, eliminating the need for anchor generation and NMS. This approach simplifies and unifies the detection pipeline but suffers from slow training convergence.

Numerous variants of DETR have been proposed to address its convergence issues by enhancing its query design. Within the decoders, each query is composed of two components: a content part and a positional one. The majority of existing research has concentrated on improving the positional part. These methods are dedicated to providing a comprehensive positional prior for each query, enabling the cross-attention module to focus on a specific region related to the target object. In contrast, the content part has been largely neglected and is typically initialized as either a zero or a learnable embedding. This offers no substantial information to the cross-attention module in the initial decoder layer.

In this paper, we focus on the content query, an aspect that has rarely been considered in previous works. We introduce a novel plug-and-play module called Self-Adaptive Content Query (SACQ) to enhance the performance of DETR's variants. Our SACQ comprises two main components: 1) globally pooled features for content query initialization, and 2) locally pooled features for further enhancement of the content query. Traditionally, the content query in a decoder is initialized with either a zero tensor or a learnable embedding, which remains static and lacks any input prior. Carion \textit{et al.} \cite{carion2020end} pointed out that the encoder in DETR already separates instances through global attention, while the decoder focuses on extremities to extract class and object boundaries. Building on this insight, we propose a self-attention pooling module (SAPM) capable of dynamically pooling features from the encoder to serve as a more effective initial content query for the decoder's first layer.

To validate our assumption, we visualize the learned sampling points with high attention weights from the first decoder layer of the original Deformable-DETR, as depicted in Figure.\ref{REF}. The visualization indicates that these points tend to either congregate in a narrow area of the predicted object or spread over the vicinity of the target. After incorporating the SAPM module, the sampling points with high weights more uniformly cover the entire predicted object, and there are significantly fewer points outside the object. This suggests that our content query supplements the content prior for each query, enabling the cross-attention module to focus better on the target object. 


The improved object queries are inclined to concentrate on target objects, causing a cluster of highly similar candidate queries generated for target objects. It poses additional challenge for the training process through conventional Hungarian one-to-one matching. Jia \textit{et al.} \cite{jia2022detrs} have pointed out that this matching strategy reduces the training efficiency of positive samples due to few queries assigned as positive samples. To alleviate this issue, we propose a straightforward solution:  merging similar predicted results from different queries into a single one before conducting set matching. The similarity of the queries is determined by the Kullback-Leibler (KL) divergence \cite{kl_div} of category predictions and the Intersection over Union (IoU) between the bounding box predictions. As demonstrated in Figure.\ref{VIS}, our SACQ module tends to produce more similar bounding boxes for target objects due to its improved initialization. By implementing the Query Aggregation (QA) strategy, we further capitalize on the benefits of SACQ by combining the outputs of these potential queries and maximizing their utility.



In summary, our technical contributions are twofold:
\begin{itemize}
    \item  We propose a novel method for content query optimization, which has been overlooked in previous works. It consists of two complementary modules: the SACQ and QA. The SACQ generates improved content query for the decoder by introducing input priors. Additionally, the QA module preserves high-quality candidates generated by the SACQ and reduces instability associated with one-to-one matching by aggregating candidate boxes. Both modules can be easily integrated into existing DETR's variants.
    \item Through extensive experiments on the COCO dataset and qualitative analysis, we validate the effectiveness of our proposed method, achieving an average improvement of over 1.0 AP across six different DETR's variants with multiple configurations.
\end{itemize}


\section{Related work}
\subsection{CNN-based Object Detection Methods}
Classical CNN-based object detectors can be divided into two categories: two-stage and one-stage methods. Two-stage methods initially generate a set of box proposals, then determine whether each proposal corresponds to an object, and finally perform bounding box regression based on the proposals. Typical methods include RCNN\cite{rcnn_2014_CVPR}, Fast-RCNN \cite{Girshick_2015_ICCV}, Faster-RCNN \cite{ren2015faster}, etc. In contrast, one-stage models directly predict the bounding box of objects based on predefined anchors or reference points. Examples of one-stage methods include SSD \cite{liu2016ssd}, YOLO series \cite{Redmon_2016_CVPR,yolov3}, and others. 

\begin{figure*}[!h]
  \centering
  \includegraphics[width=\linewidth]{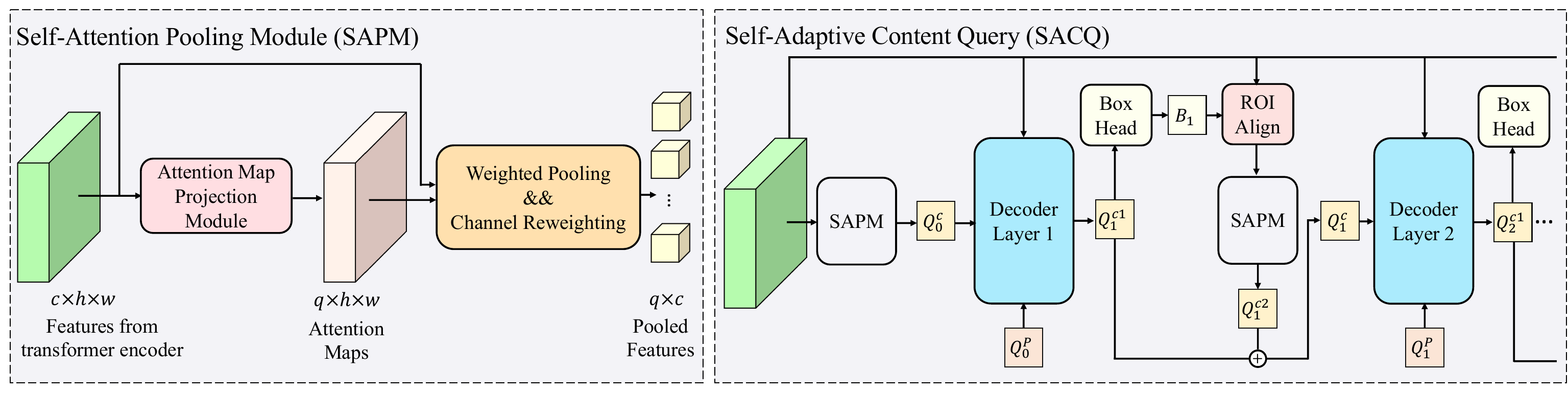}
  \caption{The left portion of the diagram depicts the structure of the proposed SAPM. Features from the transformer encoder are projected into $q$ attention maps through the attention map projection modules. For each feature from the encoder, its elements are weighted according to certain attention map in the spatial dimension and then averaged to create a spatially pooled feature. The right portion illustrates the integration of SACQ into the transformer decoder of DETR's variants. SACQ generates the initial content query for the first layer of the decoder from the features produced by the transformer encoder. Starting from the second layer of the decoder, SACQ utilizes SAPM to enhance the content query based on the previous box prediction.}
   \label{SACQ}
\end{figure*} 

\subsection{DETR and Its Variants}
DETR \cite{carion2020end} is the pioneering work that introduces transformer to object detection. Unlike previous detection methods, DETR is a true fully end-to-end detector that does not rely on hand-designed components such as anchor proposal and NMS. However, it suffers from extremely slow training convergence due to the decoder's cross-attention \cite{rethinking_trans}. Many subsequent methods have attempted to address this issue. Dai \textit{et al.} \cite{dynamic_detr} improve the encoder and decoder in DETR by incorporating dynamic attention to overcome the problems of low feature resolution and slow training convergence.
Anchor DETR\cite{yingming2021anchor} and DAB-DETR \cite{liu2022dabdetr} formulate positional queries as dynamic anchor points and anchor boxes respectively, which bridge the gap between classical anchor-based detectors and DETR-based ones. 

Some variants improve the training performance by optimizing the structure of transformer head. Sparse DETR \cite{sparse_detr} and PnP-DETR \cite{pnp_detr} address the excessive computation of the transformer network in a DETR model caused by the spatial redundancy issue of the feature map.
Deformable-DETR \cite{deformable_detr} proposes a more efficient attention module, which attends to a small set of sampling locations around a reference point as a pre-filter for prominent key elements. 

Some other works have improved upon the the query in the decoder. SAM-DETR \cite{SAM_DETR} uses query embeddings to align and reweight RoI-Aligned encoded image features and generate enhanced queries, which exhibits similarity to our query feature enhancement to some extent. Nevertheless, the main objectives of our approach and SAM-DETR are considerably different. SAM-DETR uses zero-initialized content query and primarily enhances the query after the fisrt decoder layer, whereas our SACQ aims to provide object-related content prior that was initially overlooked during query initialization in current DETR's variants. SAP-DETR \cite{SAP_DETR} assigns each query with a specific grid region and initializes the corner/center of the grid as its reference point. This method is orthogonal to our method. 
Dynamic DETR \cite{DynanmicDETR} introduces dynamic attention, which is achieved by adding additional RoI features to cross-attention into both the encoder and decoder stages to address the low feature resolution and slow training convergence problems. However, it uses learnable embedding for the initialization of the query as well, which is different from our approach. 

Li \textit{et al.} proposed DN-DETR \cite{dn_detr}, which additionally feeds ground-truth bounding boxes with noises into the transformer decoder and trains the model to reconstruct the original boxes. DINO \cite{dino} further improves denoising training by combining DN-DETR with the designs from DAB-DETR and Deformable-DETR. Mask DINO \cite{Mask_DINO} extends DINO by adding a mask prediction branch to make it support segmentation tasks. It has initialized content query by simply selecting features from encoder, but their initialization only contains information of one location which can not cover the entire target. H-DETR \cite{hdetr} and Co-DETR \cite{CO_DETR} augment additional hybrid-matching training branch, which explores more positive queries to overcome the drawbacks of one-to-one matching. Stable-DINO \cite{Liu2023DetectionTW} only utilizes positional metrics to supervise classification scores of positive examples to alleviate the instability of bipartite graph matching. Other DETR's variants proposed recently are Group-DETR \cite{group_detr}, SQR-DETR \cite{sqr_detr}, Team-DETR \cite{team_detr}, and KS-DETR \cite{KS_DETR}.

\section{Method}
\subsection{Overview}
Given an input image $I$, DETR and its variants first apply a backbone network to extract spatial features $F^B$. These features are further refined as enhanced features $F^E$ by the transformer encoder. The enhanced features, along with a default set of object queries $Q$, are then fed into the transformer decoder to identify corresponding objects. Finally, the outputs from the last layer of the decoder are used to predict labels and boxes via a prediction head.
Object queries in the transformer decoder consist of two components: a positional query $Q^p$ and a content query $Q^c$. However, in most variants of DETR, the content query is typically initialized either with zeros or with a learnable embedding. In this work, we concentrate on the content query and present a novel plug-and-play module, Self-Adaptive Content Query (SACQ), to enhance it. This is further complemented by a Query Aggregation (QA) strategy. More details will be elaborated in the following subsections.

\subsection{Self-Adaptive Content Query}

To enhance the initialization of the content query, it is essential to develop a module capable of accurately identifying and extracting object-related features from an image. Existing feature extraction methods that target specific objects, such as RoI-Align\cite{he2017mask}, necessitate the input of precise target position coordinates within the image. However, the features pooled using this method may inevitably include noise, such as the background. A promising solution to this challenge is to employ the attention mechanism to softly isolate the target, which can yield better features than those obtained via RoI-Align. This solution involves designing a module that can generate unique attention maps for each object, and using these maps to extract detailed object-specific features. These features would subsequently aid in the initialization and enhancement of the content queries. As this process does not require the input of the target's coordinates, we designate it as Self-Adaptive Content Query (SACQ), a more intuitive and autonomous method for object-related feature extraction, aiming for better content queries. 

The core of our SACQ is the Self-Attention Pooling Module (SAPM), detailed in the left part of Figure.\ref{SACQ}. The SAPM consists of three components: the Attention Map Projection (AMP) module, the Weighted Pooling (WP) module, and the Channel Reweighting (CR) module \cite{senet}. Given the input features $F \in \mathbb{R}^{c \times h \times w}$, the SAPM initially projects it into attention maps $A \in \mathbb{R}^{q \times h \times w}$ through AMP module. Here, AMP is made up of several convolutional layers. Its primary goal is to generate attention maps for each query that can focus on the corresponding target. Subsequently, the feature $F$ undergoes a weighted pooling process to derive the object-specific feature $F^P \in \mathbb{R}^{q \times c}$, which is guided by the attention map $A$ as follows:
 \begin{equation}
   F^P_i = \sum_{j=0, k=0}^{h, w} F[:,j,k]\cdot A[i, j, k], \forall i \in \{ 0,1, ..., q - 1\}.
\end{equation}
The CR module then refines the channel weights within $F^P$, thereby accentuating the distinctiveness of the extracted features. The output features can be expressed as $F^O = \sigma(\text{MLP}(F^P)) \odot F^P$, where $\sigma$ denotes the sigmoid activation function, and $\odot$ represents element-wise multiplication. 

The SAPM has been instrumental in enhancing the initialization of content queries for the first decoder layer. Furthermore, there is potential for additional optimization in subsequent layers by refining the content queries to more accurately focus on the target object. To fully capitalize on the SAPM's ability to precisely concentrate on objects, we have integrated it with RoI-Align to generate local features. This enhancement of the content queries starts from the second decoder layer and continues onwards.

The right part of Figure.\ref{SACQ} illustrates the complete SACQ module. The transformer encoder generates features $F^E$, which are initially processed by the global SAPM to produce the initial content query $Q^c_0$. This query works in conjunction with the positional query $Q^p_0$ to form a composite object query, which is used as the input for the first layer of the transformer decoder. Through multi-head cross-attention, the object query interacts with $F^E$ to produce the updated content query $Q^{c1}_1$. For the sake of brevity, feeding the positional encoding of features into the decoder is not shown in the figure. Subsequently, a box head is utilized to predict the bounding box $B_1$ for each query. These bounding boxes are then subjected to RoI-Align to extract local features specific to each predicted region. The extracted local features are then input into the local SAPM, and the resulting output $Q^{c2}_1$ is used to enhance the content query and generate the input for the next decoder layer: $Q^{c}_1 = Q^{c1}_1 + Q^{c2}_1$. The following decoder layers repeat this process of reinforcing the content query by utilizing the local SAPM with shared parameters.


\subsection{Similar Query Aggregation Strategy} 
\begin{figure}[!h]
  \centering
  \includegraphics[width=\linewidth]{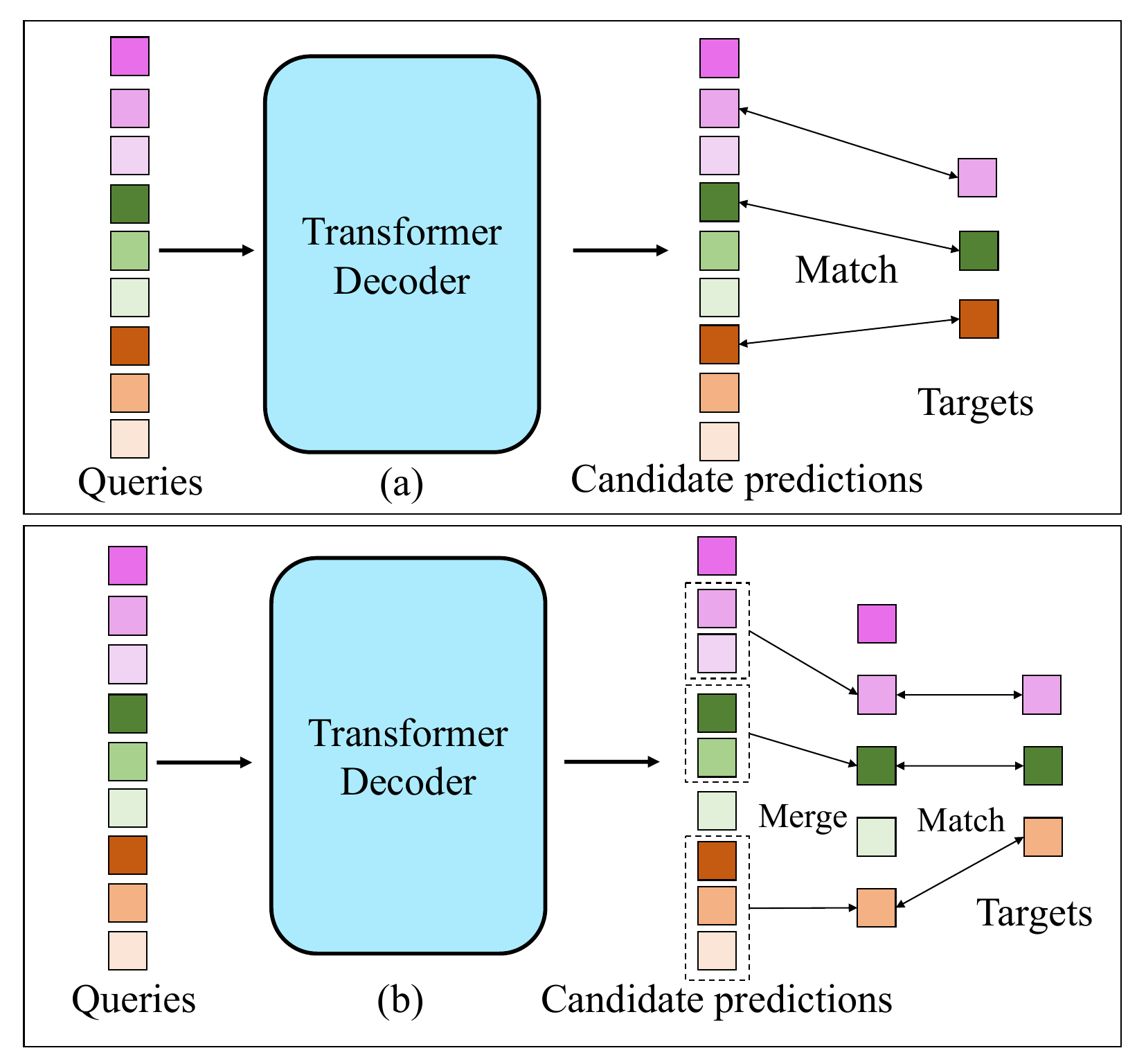}
  \caption{(a) shows the vanilla decoder of transformer. The candidate predictions generated from queries are directly matched with targets. (b) show the decoder with our query aggregation strategy. Candidate predictions are first merged according to similarity metric and then matched with targets.}
   \label{QM}
\end{figure} 

Our SACQ enhances content query through the self-attention mechanism, enabling it to produce a greater number of high-quality candidate results. Nevertheless, this improvement in candidate quality complicates the optimization process and introduces instability for the prevailing one-to-one matching mechanism. Because one-to-one matching is constrained to optimizing a single candidate per object, necessitating the suppression of any additional high-quality candidates that belong to the same object. To address this issue, we propose a method named Query Aggregation (QA) that consolidates similar predictions of distinct candidates into a unified result before set matching. This strategy not only preserves high-quality candidates but also mitigates the instability associated with one-to-one matching by aggregating candidate boxes for easier optimization. In our QA, we evaluate the similarity of predicted categories and bounding boxes between queries using Kullback-Leibler (KL) divergence and Intersection over Union (IoU), respectively. For category predictions $p_i$ and $p_j$ belonging to the $i$-th and $j$-th query, where $p_i, p_j \in \mathbb{R}^{m}$, the category similarity $S_{cls} \in \mathbb{R}^{q \times q}$ is defined as follows:
 \begin{equation}
  \begin{aligned}
     S_{cls}[i,j] &= KL(p_i || p_j) + KL(p_j || p_i) \\
      &= \sum_k^m p_i[k]log\left(\frac{p_i[k]}{p_j[k]}\right) + \sum_k^m p_j[k]log\left(\frac{p_j[k]}{p_i[k]}\right)
  \end{aligned}    
\end{equation}
For bounding box predictions $B_i$ and $B_j$ of the $i$-th and $j$-th query, box similarity $S_{box} \in \mathbb{R}^{q \times q}$ is defined by:
\begin{equation}
  \begin{aligned}
     S_{box}[i,j] &= IoU(B_i || B_j) = \frac{Area(B_i \cap B_j)}{ Area(B_i \cup B_j)}
  \end{aligned}    
\end{equation}

\begin{table*}[!h]
  \centering
  \begin{tabular}{l|c|cccccc|ccc}
    \toprule
    Method &w/ Ours &$AP$ &$AP_{50}$ &$AP_{75}$ &$AP_S$ &$AP_M$ &$AP_L$ &Epochs &FLOPs &Params\\
    \midrule
    Deformable-DETR$\dag$ & &45.4 &64.7 &49.0 &26.8 &48.3 &61.7 &50 &173 &40M \\
    Deformable-DETR$\dag$ &\Checkmark &46.9(\textbf{+1.5})	&65.4 &50.7	&29.0 &50.1	&62.5 &50 &216 &51M \\
    Deformable-DETR$\dag\dag$ & &46.2	&65.2 &50.0	&28.8 &49.2	&61.7 &50 &173 &40M \\
    Deformable-DETR$\dag\dag$ &\Checkmark &47.3(\textbf{+1.1}) &66.7 &51.2 &30.6 &50.0 &62.6 &50 &216 &51M \\
    SAM-DETR & & 41.8 &63.2 &43.9 &22.1 &45.9 &60.9 &50 &100 &58M \\
    SAM-DETR &\Checkmark &43.0(\textbf{+1.2}) &63.6 &45.9 &23.6 &47.1 &61.5 &50 &104 &62M \\
    SAP-DETR & & 43.1 &63.8 &45.4 &22.9 &47.1 &62.1 & 50  &92 &47M \\
    SAP-DETR &\Checkmark & 44.5(\textbf{+1.4}) &65.7 &47.3 &24.1 &48.4 &64.2 & 50  &95 &50M \\
    Dab-DETR	& &42.2	&63.2	&45.6	&21.8	&46.2	&61.1	&50	&94	&44M \\
    Dab-DETR &\Checkmark &43.2(\textbf{+1.0})	&63.9	&46.1	&22.5	&46.9	&61.9	&50	&97	&47M \\
    Dab-Deformable-DETR$\dag$&	&46.8	&66.0	&50.4	&29.1	&49.8	&62.3 &50	&195 &47M \\
    Dab-Deformable-DETR$\dag$&\Checkmark &47.6(\textbf{+0.8})	&65.7	&51.9	&30.0	&50.5 &62.4	&50	&239	&58M \\
    DINO & &49.0 &66.6	&53.5 &32.0	&52.3 &63.0	&12	&279 &47M \\
    DINO &\Checkmark &49.4(\textbf{+0.4})  &66.9 	&53.9  & 31.8	&52.3  &64.6  &12 &323 &58M \\
    DN-DETR	& &41.1	&61.7	&43.5	&20.6	&44.8	&59.6	&12	&94	&44M \\
    DN-DETR &\Checkmark &42.4(\textbf{+1.3}) &63.0 &45.1 &21.7	&46.3 &61.2 &12	&97	&47M \\
    DN-Deformable-DETR$\dag$& &43.4 &61.9	&47.2 &24.8	&46.8 &59.4	&12	&195 &48M \\
    DN-Deformable-DETR$\dag$&\Checkmark &44.5(\textbf{+1.1}) 	&63.3 &47.8	&27.5 &48.0	&58.9 &12 &239 &59M \\
    \bottomrule
  \end{tabular}
 \caption{Experimental results based on Deformable-DETR, SAM-DETR, SAP-DETR, Dab-DETR, DINO and DN-DETR on COCO validation set. $\dag$ indicates that Deformable-DETR uses iterative bounding box refinement mechanism. $\dag\dag$ stands for two-stage Deformable-DETR, which utilizes generated region proposals of the first stage as object queries for further refinement.}
  \label{main_res}
\end{table*}

Here, $q$ represents the number of queries, and $m$ is the total number of object categories. We establish two thresholds to determine which queries to merge: a category similarity threshold $t_c$, and a box similarity threshold $t_b$. The criteria for merging are $S_{cls} < t_c$ and $S_{box} > t_b$. For a set of $n$ queries identified for merging $Q_i$, where $i\in M$, the merged result is calculated by averaging the predictions: $p = \frac{1}{n}\sum_{i \in M} p_i$, $B=\frac{1}{n}\sum_{i \in M} B_i$. Figure.\ref{QM} illustrates the distinction between a transformer decoder employing our query aggregation strategy and a vanilla transformer decoder.

\section{Experiments}

\subsection{Setup}
\paragraph{\textbf{Dataset}} We conduct the experiments on the well-known COCO 2017 object detection dataset \cite{coco}, which contains about 118K training images and 5K validation images. Following the common practice in detection methods, we report the standard mean average precision (AP) result on the validation set under different bounding box IoU thresholds with different object scales.

\paragraph{\textbf{Implemention Details}} We test the effectiveness of our method on six DETR's variants: Deformable-DETR, SAM-DETR, SAP-DETR, DAB-DETR, DN-DETR, and DINO. They comprise a backbone network, multiple transformer encoder layers, and decoder layers. For a fair comparison, we uniformly adopt ResNet-50 \cite{resnet} model pre-trained on ImageNet-1K \cite{imagenet} as the backbone for each variant. We follow the original hyper-parameters setting of corresponding baseline methods. For detailed network structure of SAPM please refer to Appendices.A.1. The category and box similarity thresholds  are set as: $t_c=3 \times 10^{-7}, t_b=0.9$. The output size of RoI-Align used in content query enhancement is $7 \times 7$. We use $2$ images per GPU on an 8-(A100)GPU machine for training, with a total batch size of 16. AdamW \cite{adamw} is used for optimizing with $\beta_1=0.9$, $\beta _2=0.999$, and weight decay $10^{-4}$. The learning rates for the backbone network and other modules are set to $10^{-5}$ and $10^{-4}$, respectively. For fast converging variants (DN-DETR and DINO), we train models for $12$ epochs and drop the learning rate by $0.1$ after $11$ epochs. For Deformable-DETR and DAB-DETR, we train models for 50 epochs and drop the learning rate by $0.1$ after $40$ epochs. For the loss function, we use the L1 loss and GIOU \cite{giou} loss for bounding box regression and focal loss \cite{retina_2019_ICCV} with $\alpha = 0.25$, $\gamma = 2$ for object classification. Following the training setting in DETR's variants, we add auxiliary losses after each decoder layer. We use the same loss coefficients as each baseline method, that is, $2.0$ for classification loss, $5.0$ for L1 loss, and $2.0$ for GIOU loss.

\subsection{Main Results}

Table \ref{main_res} presents our main experimental results. All models are evaluated on the COCO 2017 validation set for fairness. Our method consistently enhances the performance of all methods. For Deformable-DETR, our approach achieves AP gains of $1.5$ ($45.4$ vs. $46.9$) and $1.1$ ($46.2$ vs. $47.3$) under iterative bounding box refinement and two-stage settings, respectively. DAB-DETR and DAB-Deformable-DETR improve the positional aspect of the query, and our method further enhances performance with AP gains of $1.0$ ($42.2$ vs. $43.2$) and $0.8$ ($46.8$ vs. $47.6$), respectively. This indicates that our optimization of the content query is orthogonal to the position query. For SAM-DETR and SAP-DET, our approach results in AP gains of $1.2$ ($41.8$ vs. $43.0$)and $1.4$ ($43.1$ vs. $44.5$) respectively. DN-DETR, which introduces a query denoising task to help stabilize bipartite graph matching and accelerate training convergence, also benefits from our method with a $1.3$ AP improvement ($41.1$ vs. $42.4$) under a 12-epoch training schedule. For state-of-the-art method DINO, we obtains $0.4$ ($49.0$ vs. $49.4$) AP improvement. Although the gains for DINO are not yet significant, our joint optimization of both the content query and matching strategy has illuminated a new direction for DETR-based detection methods. These two modules are closely related in a non-trivial way. Our current solution is effective and has great potential for further improvement, which we leave for future research. For more results of Swin Transformer backbone \cite{Swin} please refer to Appendices.B.1.

\subsection{Ablations}
\begin{table}[!h]
  \centering
  \setlength{\tabcolsep}{2.5pt}
  \begin{tabular}{ccc|cccccc}
    \toprule
    \multicolumn{2}{c}{SACQ}  & \multirow{2}{*}{QA} & \multirow{2}{*}{$AP$} &\multirow{2}{*}{$AP_{50}$} &\multirow{2}{*}{$AP_{75}$} &\multirow{2}{*}{$AP_S$} &\multirow{2}{*}{$AP_M$} &\multirow{2}{*}{$AP_L$} \\
    Global & Local &  & & & & & & \\
    \midrule
    \multicolumn{3}{c|}{baseline}&45.4 &64.7 &49.0 &26.8 &48.3 &61.7\\
    \Checkmark & & &46.2 &65.2 &49.7 &28.7 &49.2 &61.9 \\
    \Checkmark &\Checkmark &  &46.6 &65.2 &50.4&28.9 &49.5 &62.3 \\
    \Checkmark &\Checkmark &\Checkmark  &46.9	&65.4 &50.7	&29.0 &50.1	&62.5 \\
    \bottomrule
  \end{tabular}
  \caption{Ablation results for our method. All models are tested over Deformable-DETR with iterative bounding box refinement baseline.}
  \label{ablation1}
\end{table}

\begin{figure*}[!h]
  \centering
  \includegraphics[width=\linewidth]{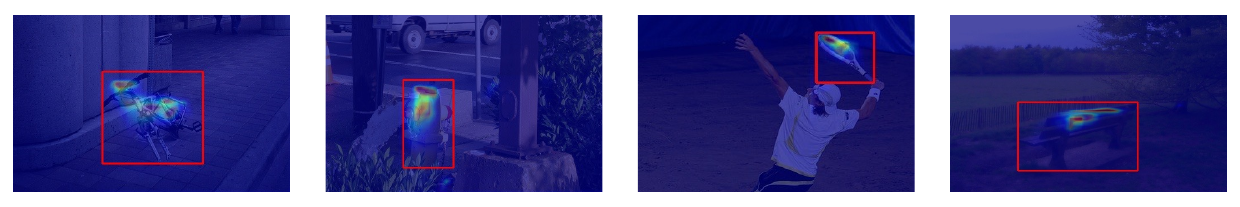}
  \caption{The attention maps from the SACQ module are visualized to correspond with detected objects, each encased within a red bounding box. These maps exhibit a well focus on the predicted object, indicating their efficacy in extracting features that are relevant to the target. The ability to precisely concentrate on specific object confirms that generated features are suitable for the initialization of content queries.}
  \label{VAM}
\end{figure*} 

We conduct a set of ablation studies on Deformable-DETR with iterative bounding box refinement baseline to verify the effectiveness of each component in our method. The results in Table \ref{ablation1} show that all components contribute to performance improvement. SACQ-Global means we only adopt one SAPM to pool global features from the encoder to initialize content queries. SACQ-Local denotes locally pooled features are used to enhance the content queries after the first decoder layer. QA stands for our similarity query aggregation strategy, which merges similar predicted results from different queries into the same one. The results show that content query initialized with globally pooled feature has the most noticeable performance improvement.

\begin{table}[!h]
  \centering
  \setlength{\tabcolsep}{3pt}
  \begin{tabular}{c|cccccc}
    \toprule
    with CR module &{$AP$} &$AP_{50}$ &$AP_{75}$ &$AP_S$ &$AP_M$ &$AP_L$ \\
    \midrule
               &46.3 &65.1 &49.8 &28.7 &49.2 &62.2 \\
    \Checkmark &46.6 &65.2 &50.4 &28.9 &49.5 &62.3 \\
    \bottomrule
  \end{tabular}
  \caption{The influence of CR module for SACQ.}
  \label{ablation2}
\end{table}

We analyze the influence of the channel weighting module of SACQ, as shown in Table \ref{ablation2}. The results indicate that adding the CR module improves performance to some extent. We argue that the CR module can make each content query more specialized and respond to different inputs in a highly object-specific manner. 


Furthermore, we investigate the impact of varying thresholds on our query aggregation strategy. We set the category threshold at a low value to guarantee that queries with the same categories are merged. This does not have a significant impact on performance outcomes. However, the performance is highly sensitive to the bounding box Intersection over Union (IoU) threshold. We observe a performance decline when the box IoU threshold is too small. As shown in Table \ref{ablation4}, we present the results obtained using various box IoU thresholds. When the threshold is set at $0.7$, the performance declines to an AP of $45.3$, which falls below the baseline without query aggregation. This decrease can be attributed to the negative impact on the merging of objects that do not significantly overlap with one another. For more ablations, please refer to Appendices B.2 and B.3.

\begin{table}[!h]
  \centering
  \setlength{\tabcolsep}{2.1pt}
  \begin{tabular}{c|cccccc}
    \toprule
    Box IoU threshold $t_b$ &{$AP$} &$AP_{50}$ &$AP_{75}$ &$AP_S$ &$AP_M$ &$AP_L$ \\
    \midrule
    0.9 &46.9 &65.4 &50.7 &29.0 &50.1 &62.5 \\
    0.8 &46.7 &65.2 &50.5 &28.9 &49.6 &62.3 \\
    0.7 &45.3 &63.4 &48.4 &27.6 &48.5 &61.0 \\
    \bottomrule
  \end{tabular}
  \caption{The influence of different box IoU thresholds in QA.}
  \label{ablation4}
\end{table}

\subsection{Discussions}
\paragraph{\textbf{What do attention maps of SACQ learn?}}  

Our comprehensive experiments conducted across various baselines have confirmed the effectiveness of our SACQ. To provide a clear understanding of its self-attention mechanism, we have visualized the global pooling attention maps in the form of heatmaps. As depicted in Figure \ref{VAM}, each attention map within the SACQ module accurately concentrates on the related object (indicated by the red bounding box, which represents the predicted object of the corresponding query). For queries with low prediction scores, the attention maps exhibit a more uniform distribution, suggesting a less focused attention. The capacity to precisely focus on specific objects verifies that the features generated are appropriate for initiating content queries. This initiation results in a superior content prior for cross-attention calculations in the initial decoder layer, thereby improving the cross-attention mechanism's precision in targeting the desired objects. For additional visualizations, please see Appendix B.4.

\paragraph{\textbf{Can SACQ be replaced by ROI-Aligned features?}} 

The ROI-aligned results on encoded feature maps can simply serve as an option for content query initialization. However, it necessitates an additional module to generate ROIs for most DETR's variants (except two-stage Deformable-DERT). This contradicts one of the key advantages of DETR's variants, namely the elimination of anchor or proposal generation. Additionally, we conduct the experiment of using ROI-aligned features as content query initialization, where the ROIs are from the first stage of two-stage Deformable-DERT. Compared to the original two-stage Deformable-DERT, the performance deteriorates 1.1 points (45.1 vs. 46.2). The primary reasons are twofold:  1) the bounding boxes predicted from the first stage is of low quality, as indicated by DINO's author; 2) the obtained features using ROIs contain irrelevant contents as the objects may not perfectly fit the target boxes, making the features ambiguous and insufficient for content query initialization. In contrast, SACQ can accurately focus on target objects through SAPM module (refer to Figure.\ref{VAM}).

\paragraph{\textbf{How does QA cooperate with SACQ?}}

\begin{figure}[!h]
  \centering
  \includegraphics[width=\linewidth]{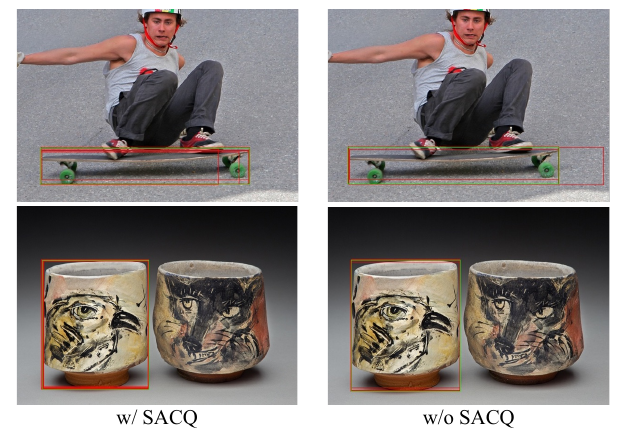}
  \caption{Visualization of activated query's bounding box (green boxes) and its highly overlapped (IoU $> 0.8$) bounding boxes (red boxes) from queries with suppressed low scores. Improved content query initialization from SACQ generates more potential queries with similar bboxes, which can be further addressed by QA.}
   \label{VIS}
\end{figure}

With improved initialization, SACQ is capable of producing a greater number of high-quality candidate bounding boxes for a target object, as illustrated in the left part of Figure.\ref{VIS}. The traditional one-to-one matching approach would assign a high target score to only one of these queries, resulting in the suppression and underutilization of the remaining queries. Additionally, the presence of more high-quality candidates can further destabilize the optimization process. For example, candidates A and B both meet the matching criteria for a target object. During a specific training iteration, candidate A might be optimized while candidate B is suppressed, and vice versa in another iteration. This fluctuation aggravates the instability of the optimization process and makes it more difficult to achieve convergence. Our Query Aggregation (QA) module is designed to address this issue by merging the outputs of these high-quality candidates, thereby eliminating the need to suppress any additional high-quality candidates corresponding to the same object.

\begin{figure}[!h]
  \centering
  \includegraphics[width=\linewidth]{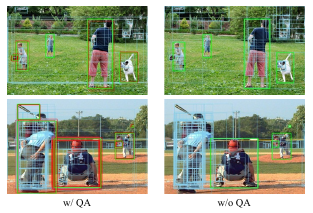}
  \caption{Visualization of all predicted boxes of queries in Deformable-DETR w/ and w/o QA. The green and red boxes are the predicted objects of merged queries and corresponding queries before merging, respectively. And the blue boxes are from queries with low predicted scores ($< 0.5$)}
   \label{VSQ}
\end{figure}

\paragraph{\textbf{What does the object predicted by the merged query look like?}} 
As previously discussed, our query aggregation strategy combines similar predictions from different high-quality candidates into a single prediction. In the validation set, the maximum number of merging operations is 169, while the minimum is 1, indicating instances where no merging occurred. Figure.\ref{VSQ} illustrates the predicted bounding boxes after merging, alongside the original prediction from each query. The green bounding boxes represent predictions from merged queries with scores above 0.5, while the red boxes denote predictions from queries before merging. The blue boxes represent the predictions from queries with scores below 0.5. The results demonstrate that our strategy can increase confidence in object prediction by merging high-quality candidates and maximizing their utility. For example, in Figure.\ref{VSQ}, the score for the person on the left in the second-row image without query aggregation is below 0.5. However, with query aggregation, the corresponding prediction for the same person exceeds a score of 0.5. This highlights the effectiveness of our aggregation approach in improving the reliability of object detection.

\section{Conclusion}
In this paper, we introduce a novel plug-and-play method that enhances the performance of DETR's variants. Our approach incorporates a Self-Adaptive Content Query (SACQ) module and a Query Aggregation (QA) strategy. The SACQ module improves the content aspect of the query in DETR's variants by offering better initialization and step-by-step enhancement. The QA strategy, on the other hand, preserves the high-quality candidates generated by SACQ and reduces the instability associated with one-to-one matching by merging similar candidate boxes. This further complements the SACQ module. We have conducted extensive experiments on six different baseline methods with multiple configurations to validate the efficacy of our approach.

\appendix
\section{Implemention Details}
\subsection{Detailed Network Structure of SAPM}

\begin{figure}[!h]
    \begin{minipage}[t]{0.5\linewidth}
        \centering
        \includegraphics[width=1.0 in]{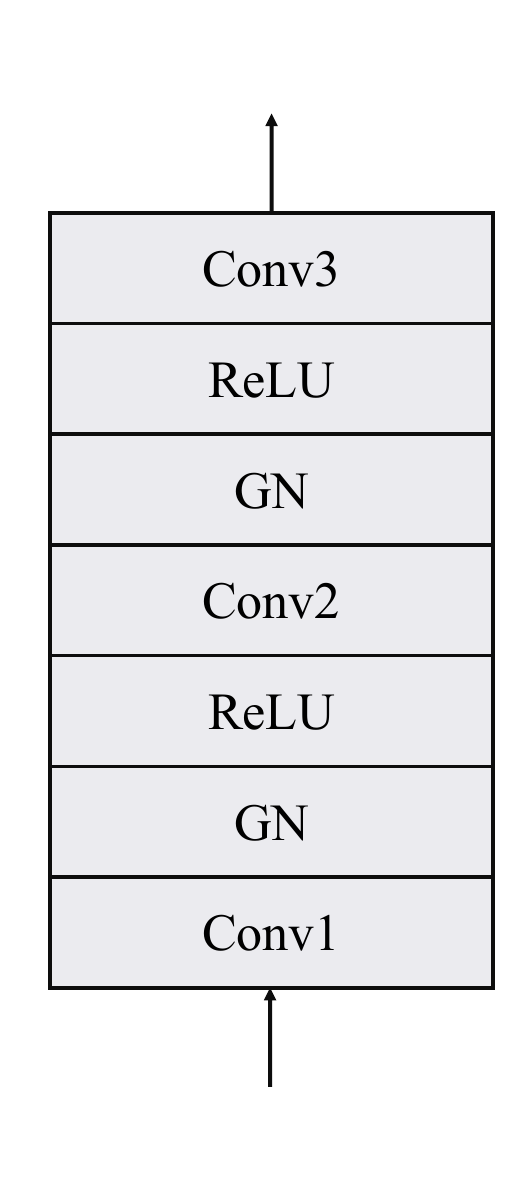}
        \centerline{(a) Attention Map Projection}
    \end{minipage}%
    \begin{minipage}[t]{0.5\linewidth}
        \centering
        \includegraphics[width=1.0 in]{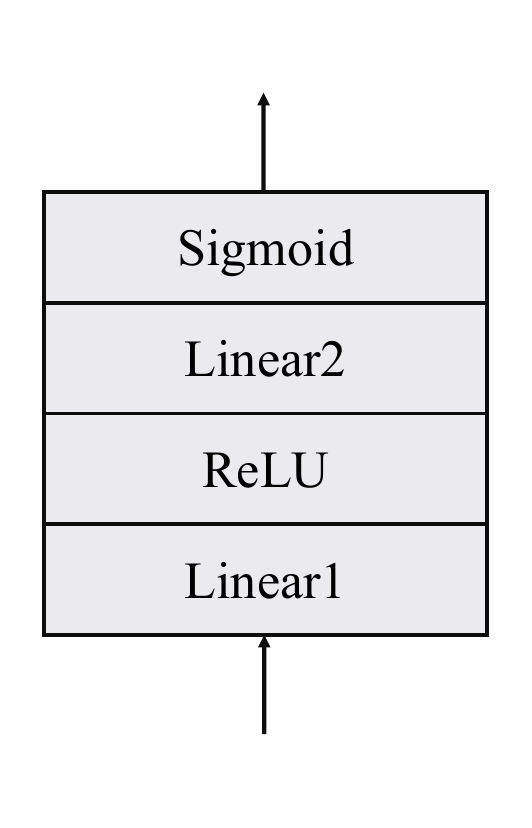}
        \centerline{(b) Channel Reweghting}
    \end{minipage}
    \caption{Detailed Architecture of SAPM} 
    \label{sapm}
\end{figure}

In our experiments, AMP module is made up of three convolutional layers, which is illustrated in Fig.\ref{sapm}(a). Features from transformer encoder have 256 channels. Hence, the kernel shape of Conv1 is: $256 \times 256 \times 5 \times 5$. The kernel shape of Conv2 is: $256 \times 256 \times 3 \times 3$. And the kernel shape of Conv3 is: $q \times 256 \times 3 \times 3$. $q$ is the number of queries in DETR's variants. ($q=900$ for DINO and $q=300$ for other variants). The number of groups in GN is 32. We use softmax function to normalize the attention map on spatial dimension: 
 \begin{equation}
  \begin{aligned}
     A[i, j, k] = \frac{\exp(F[i, j, k] *  \tau)}{\sum_{j, k}^{h, w}\exp(F[i, j, k] *  \tau)}
  \end{aligned}    
\end{equation}
Here, $F \in \mathbb{R}^{q \times h \times w}$ is the output feature of Conv3. We set $\tau=1.2$ in our experiments. For Deformable-DETR and DINO, which have multi-scale features, AMPs with independent parameters are used to generate pooled features for each scale. Furthermore, the pooled features of different scales are averaged as the final feature: 
 \begin{equation}
  \begin{aligned}
     F^P = \frac{1}{N} \sum_i^N F^P_{(i)}
  \end{aligned}    
\end{equation}
Here, $F^P_{(i)}$ is the i-th pooled feature, and $N$ is the total number of scales of features. As is shown in Fig.\ref{sapm}(b), CR module is made up of two linear layers, the shape of each layer's weight is: $256 \times 256$.

\subsection{About Positional Query in Our Method}
We would like to clarify that we did not change the positional query of each DERT's variant. In the case of Dab-DETR, the positional query is an embedding of the anchor box, which is refined layer by layer. For the other variants, the positional query remains the same across different layers. We will highlight this point in a revision of the paper.

\subsection{Details of SAPM in Two-stage Deformable DETR}
 In the original two-stage deformable DETR, the top-k inactive bounding box coordinates are selected from the first stage to create a set of positional embeddings consisting of k d-dimensional vectors. These positional embeddings are then processed through a layer of fully connected (FC) followed by layer normalization (LN), resulting in 2d-dimensional vectors. These vectors are subsequently divided into two separate d-dimensional vectors: the positional query and content query. Thus, the original two-stage deformable DETR's query only contains location information and lacks the content information of the object. In our approach, we replace the content query with the query generated by SAPM, just like replacing a zero-initialized or learnable content query in other variants. In the case of one-stage deformable DETR, both the positional query and content query are initialized by learned embedding, we also only substitute the  initialization of content query.  It should be emphasized that we do not change the positional query of each DERT's variant.

\section{More Experiments}
\subsection{Results of Swin-tiny Backbone}
We also provide experiments results of our method based on swin-tiny backbone network. Table \ref{main_res_swin} shows the performance gains of our method in different DETR's variants. For DN-DETR and DINO, we train models for $12$ epochs and drop the learning rate by $0.1$ after $11$ epochs. For others, we train models for 50 epochs and drop the learning rate by $0.1$ after $40$ epochs.

\begin{table}[!h]
  \centering
  \setlength{\tabcolsep}{0.1pt}
  \begin{tabular}{l|c|cccccc}
    \toprule
    Method &w/ Ours &$AP$ &$AP_{50}$ &$AP_{75}$ &$AP_S$ &$AP_M$ &$AP_L$ \\
    \midrule
    Deform.-DETR$\dag$ & &49.7 &68.6 &54.1 &32.8 &52.2 &65.8  \\
    Deform.-DETR$\dag$ &\Checkmark &50.8 &69.7 &55.3 &33.3 &53.7 &66.4\\
    Deform.-DETR$\dag\dag$ & &50.9	&69.6 &55.5	&33.6 &53.4	&66.6\\
    Deform.-DETR$\dag\dag$ &\Checkmark &51.7 &70.3 &56.6 &34.6 &54.3 &67.3 \\
    SAM-DETR & & 45.1 &63.2 &43.9 &25.8 &49.2 &64.5  \\
    SAM-DETR &\Checkmark &45.8 &64.9 &46.1 &26.6 &50.1 &65.5  \\
    SAP-DETR & & 45.9 &65.5 &47.5 &25.5 &49.8 &65.1 \\
    SAP-DETR &\Checkmark & 46.5 &66.2 &48.3 &26.1 &50.4 &66.0 \\
    Dab-DETR	& &44.3	&65.8	&47.1	&24.3	&48.0	&64.4	\\
    Dab-DETR &\Checkmark &45.2	&66.9	&48.0	&24.9	&48.9	&65.7\\
    DINO & &51.3 &69.6	&56.2 &34.4	&53.8 &66.5	 \\
    DINO &\Checkmark &51.8 &70.4 &56.7  & 34.6	&54.4  &67.5  \\
    DN-DETR	& &43.6	&63.5	&45.8	&22.9	&46.9	&62.1	\\
    DN-DETR &\Checkmark &44.7 &64.7 &46.9 &23.7	&47.5 &63.4 \\
    \bottomrule
  \end{tabular}
 \caption{Experimental results based on swin-tiny backbone on COCO validation set. $\dag$ indicates that Deformable-DETR uses iterative bounding box refinement mechanism. $\dag\dag$ stands for two-stage Deformable-DETR, which utilizes generated region proposals of the first stage as object queries for further refinement.}
  \label{main_res_swin}
\end{table}

\subsection{Different normalization functions}
Table \ref{ablation3} shows the different results between sigmoid and softmax normalization for the attention map used in SACQ. Softmax normalization achieve better performance due to the attention maps can make content queries focus on objects better.

\begin{table}[!h]
  \centering
  \setlength{\tabcolsep}{3.2pt}
  \begin{tabular}{c|cccccc}
    \toprule
    Norm. function&{$AP$} &$AP_{50}$ &$AP_{75}$ &$AP_S$ &$AP_M$ &$AP_L$ \\
    \midrule
    sigmoid &46.0 &64.8 &49.6 &28.0 &49.1 &61.8 \\
    softmax &46.6 &65.2 &50.4 &28.9 &49.5 &62.3 \\
    \bottomrule
  \end{tabular}
  \caption{The influence of different normalization functions of attention map in SACQ.}
  \label{ablation3}
\end{table}

\subsection{Influence of The Number of Conv. Layer}
We also show the influence of the number of convolutional layers of the attention map projection module in our SACQ module. As shown in Table \ref{ablation4}, the results demonstrate that adding more convolutional layers improves the performance, but the performance improvement becomes marginal as the number of layers increases. Therefore, we use $3$ convolutional layers in the attention map projection module in our experiment.

\begin{table}[!h]
  \centering
  \setlength{\tabcolsep}{4.5pt}
  \begin{tabular}{c|cccccc}
    \toprule
    Conv. layers&{$AP$} &$AP_{50}$ &$AP_{75}$ &$AP_S$ &$AP_M$ &$AP_L$ \\
    \midrule
    1 &46.3 &64.4 &49.9 &28.9 &49.1 &62.0 \\
    2 &46.4 &64.6 &50.0 &28.8 &49.3 &61.7 \\
    3 &46.6 &65.2 &50.4 &28.9 &49.5 &62.3 \\
    4 &46.7 &65.1 &50.3 &29.1 &49.8 &62.5 \\
    5 &46.7 &64.9 &50.4 &29.3 &49.8 &62.3 \\
    \bottomrule
  \end{tabular}
  \caption{The influence of the number of convolutional layers of the attention map projection module in SACQ.}
  \label{ablation4}
\end{table}

\subsection{Training and Inference Time}
We conduct experiments on A100 GPUs.  The total training time of original two-stage deformable DETR is 37.3 hours (50 epochs, 8 GPUs). The training time increases to 44.5 hours after adding SAPM and QA. The inference FPS are 25.1 and 21.7 for original and our approach, respectively. For DINO, our method increases 3.1 hours(12 epochs,  16.3 hours vs. 19.4 hours). The inference FPS drops 2.7 ( 18.5 vs. 15.7).

\subsection{More Visualizations of Attention Maps of SACQ}
  We visualize the global pooling attention maps in corresponding to features at different scales in the form of heat maps. As depicted in Fig.\ref{VAM}, each attention map in SACQ module focuses on its related object (predicted object of corresponding query which is drawn as red bounding box). Furthermore, the attention maps corresponding to high-resolution (low-level) features have smaller highlight areas than those corresponding to low-resolution (high-level) features.  That means attention maps of high-resolution features focus on the Local salient features of objects. And attention maps of low-resolution features provide more global information for objects in the pooling procedure. The pooled features through these attention maps, which focus on a local region corresponding to a target object can provide better content prior for cross-attention computing in the first decoder layer. And the content prior will make the cross-attention in decoder focus on a target object better.

\begin{figure*}[!h]
  \centering
  \includegraphics[width=\linewidth]{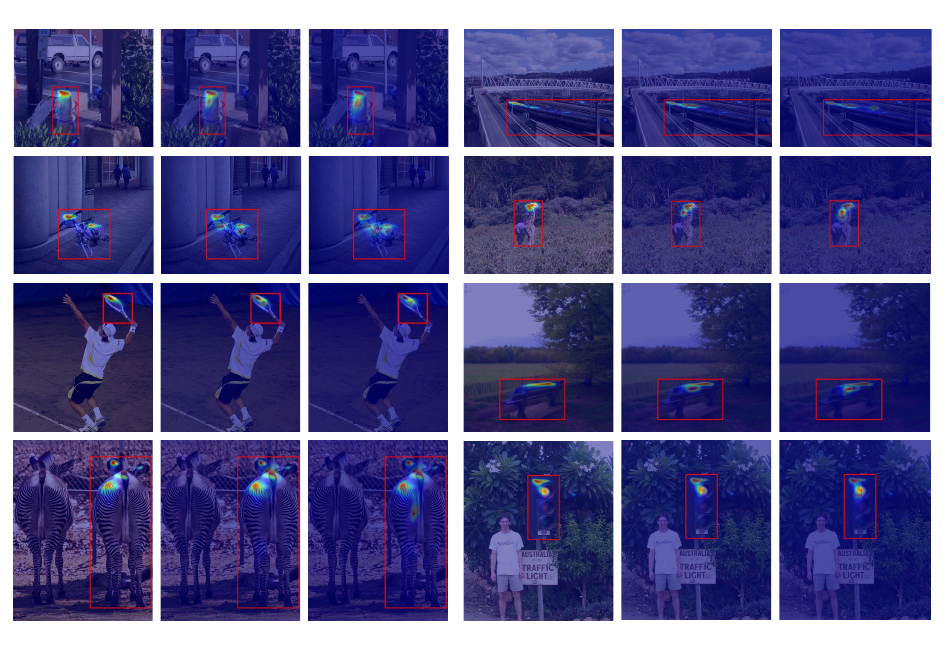}
  \caption{Visualization of attention maps of SACQ for detected objects. For each object, we show three attention maps corresponding to different-scale features of Deformable-DETR. From left to right, the first attention map corresponds to the resolution of $C_3$-stage features of the backbone network. And the next two correspond accordingly to $C_4$-stage and $C_5$-stage features.}
  \label{VAM}
\end{figure*}

\bibliographystyle{named}
\bibliography{ICQ_SQA}

\end{document}